\documentclass{article}

\usepackage{microtype}
\usepackage{graphicx}
\usepackage{subfigure}
\usepackage{booktabs} 

\usepackage{hyperref}

\usepackage[accepted]{mlsys2024}

\mlsystitlerunning{UniDM: A Unified Framework for Data Manipulation with Large Language Models}

\usepackage{tabulary,multirow,overpic}
\usepackage{amssymb}
\usepackage{algorithm}
\usepackage{algorithmic}
\usepackage{hyperref}
\usepackage{fancybox}
\usepackage[titletoc]{appendix}
\usepackage{xspace}
\usepackage{soul} 
\usepackage{color}
\usepackage{xcolor}
\usepackage{amsmath}
\usepackage{pythonhighlight}
\definecolor{shadecolor}{rgb}{0.92,0.92,0.92}  
\usepackage{tcolorbox} 

\usepackage{array}
\makeatletter
\newcommand{\thickhline}{%
    \noalign {\ifnum 0=`}\fi \hrule height 1.2pt
    \futurelet \reserved@a \@xhline
}

\newcommand{\sstitle}[1]{\vspace{0.6ex}\noindent\underline{\textbf{#1}}}

\definecolor{mygray}{RGB}{230,230,240}
\definecolor{myblue}{RGB}{175, 238, 235}

\newcommand{\sys}{UniDM\xspace}
\newcommand{\eat}[1]{}

\newcommand{\squishlist}{
	\begin{list}{$\bullet$}{
		\setlength{\itemsep}{0pt}
		\setlength{\parsep}{3pt}
		\setlength{\topsep}{3pt}
		\setlength{\partopsep}{0pt}
		\setlength{\leftmargin}{1.0em}
		\setlength{\labelwidth}{1em}
		\setlength{\labelsep}{0.5em}
   }
}

\newcommand{\squishenum}{
	
	\begin{list}{\usecounter{scount}}{
		\setlength{\itemsep}{0pt}
		\setlength{\parsep}{3pt}
		\setlength{\topsep}{3pt}
		\setlength{\partopsep}{0pt}
		\setlength{\leftmargin}{1.2em}
		\setlength{\labelwidth}{1em}
		\setlength{\labelsep}{0.5em}
	}
}

\newcommand{\squishend}{
	\end{list}
}
\begin{document}

\twocolumn[
\mlsystitle{UniDM: A Unified Framework for Data Manipulation with Large Language Models}



\mlsyssetsymbol{equal}{*}

\begin{mlsysauthorlist}
\mlsysauthor{Yichen Qian}{equal,ali}
\mlsysauthor{Yongyi He}{equal,ali,ustc}
\mlsysauthor{Rong Zhu}{ali}
\mlsysauthor{Jintao Huang}{ali,ustc}
\mlsysauthor{Zhijian Ma}{ali}
\mlsysauthor{Haibin Wang}{ali}
\mlsysauthor{Yaohua Wang}{ali}
\mlsysauthor{Xiuyu Sun}{ali}
\mlsysauthor{Defu Lian}{ustc}
\mlsysauthor{Bolin Ding}{ali}
\mlsysauthor{Jingren Zhou}{ali}
\end{mlsysauthorlist}

\mlsysaffiliation{ali}{Alibaba Group}
\mlsysaffiliation{ustc}{University of Science and Technology of China}

\mlsyscorrespondingauthor{Rong Zhu}{red.zr@alibaba-inc.com}
\mlsyscorrespondingauthor{Defu Lian}{liandefu@ustc.edu.cn}
\mlsyscorrespondingauthor{Bolin Ding}{bolin.ding@alibaba-inc.com}
\mlsyscorrespondingauthor{Jingren Zhou}{jingren.zhou@alibaba-inc.com}


\mlsyskeywords{Data Manipulation, Large Language Models, Unified Framework, Data Retrieval}

\vskip 0.3in

\begin{abstract}
Designing effective data manipulation methods is a long standing problem in data lakes.
Traditional methods, which rely on rules or machine learning models, require extensive human efforts on training data collection and tuning models.
Recent methods apply Large Language Models (LLMs) to resolve multiple data manipulation tasks. They exhibit bright benefits in terms of performance but still require customized designs to fit each specific task. This is very costly and can not catch up with the requirements of big data lake platforms. 
In this paper, inspired by the cross-task generality of LLMs on NLP tasks, we pave the \emph{first} step to design an \emph{automatic} and 
\emph{general} solution to tackle with data manipulation tasks. We propose \sys, a \underline{uni}fied framework which establishes a new paradigm to process \underline{d}ata \underline{m}anipulation tasks using LLMs. \sys formalizes a number of data manipulation tasks in a unified form and abstracts three main general steps to solve each task. 
We develop an automatic context retrieval to allow the LLMs to retrieve data from data lakes, potentially containing evidence and factual information. For each step, we design effective prompts to guide LLMs to produce high quality results.
By our comprehensive evaluation on a variety of benchmarks, our \sys exhibits great generality and state-of-the-art performance on a wide variety of data manipulation tasks.
\end{abstract}
]



\printAffiliationsAndNotice{\mlsysEqualContribution}

\section{Introduction}
\label{sec:intro}


Data lake is a general system to store vast amounts of data with heterogeneous schemas and structures. It provides an efficient interface that allows users to manage and manipulate data with various kinds of tools. Users could flexibly define different workflows to clean, integrate, interpret and analyze data according to their applications~\cite{nargesian2019data,ouellette2021ronin}. This advantage facilitates users' customized demands on data processing, but also brings remarkable shortcomings. For any new application, the corresponding data processing workflow needs to be redesigned and tuned by experts from scratch. This is very costly and can not catch up with the new applications which may occur every day in big data lake platforms~\cite{hai2021data}.

Literature works have devoted considerable research efforts to designing \emph{automatic} and \emph{general} methods that are applicable to different data manipulation tasks in data lakes. Traditional rule-based methods~\cite{dalvi2013optimal,singh2017synthesizing,chu2013holistic,chu2015katara,dallachiesa2013nadeef,mayfield2010eracer,jin2020auto} require specialized model construction and rule tuning for each data task, which are not automatic enough. Recent works apply machine learning~\cite{bilenko2003adaptive,konda2016magellan,heidari2019holodetect,DataWig,li2021opengauss,li2021cleanml,wu2021unified,alserafi2019keeping}, especially deep learning techniques~\cite{mudgal2018deep,ebraheem2018distributed,zhao2019auto,deng2022turl}, to learn the adaptive solution for each task. However, their performance heavily relies on the quality of the trained models, which require a large amount of labeled training data and specific domain knowledge related to each task. 

\begin{figure*}[!t]
  \centering
  \includegraphics[width=\linewidth]{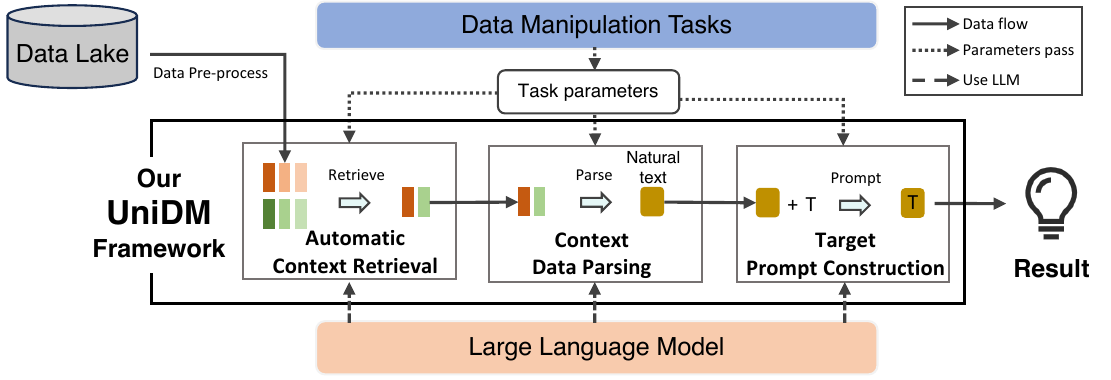}
  \vspace{-1em}
  \caption{An overview of our \sys framework.}
  \label{fig:overview}
  \vspace{-1.em}
\end{figure*}

In recent time, Large Language Models (LLMs), such as BERT~\cite{devlin2018bert}, GPT-3~\cite{brown2020language}, and LaMDA~\cite{thoppilan2022lamda}, have shown incredible performance on a broad set of downstream tasks~\cite{zhou2023comprehensive,liang2022holistic}. 
LLMs are typically deep neural networks with transformer architecture~\cite{vaswani2017attention}. They are pre-trained on enormous corpora of text to learn universal world knowledge. 
On NLP tasks, LLMs have shown remarkable cross-task generality. This is because the NLP field has accumulated decades of experience in designing a standard paradigm to unify and solve different NLP tasks~\cite{radford2019language,brown2020language}. Whereas, for data manipulation tasks, the relevant experience is almost blank, which makes this problem to be extensively challenging. To resolve it, we need to answer the following two key questions:

1) \emph{How to design a framework to elegantly unify different data manipulation tasks?} This framework should be general enough to subsume common and new tasks occurred in data lake applications and easy to bring LLMs into the solution.

2) \emph{How to design a general solution under this unified framework?} This solution should contain abstract procedures that are adaptive to different tasks and at the same time, maximize the effectiveness of LLMs.

\sstitle{Our Contributions.}
In this paper, we pave the first step towards resolving this problem. We propose \sys, a unified solution that is verified to attain state-of-the-art performance on a variety of data manipulation tasks on data lakes. Specifically, we make the following contributions:

1) \textbf{We propose a unified framework to describe data manipulation tasks.} We formalize a data manipulation task $T$ as a function $F_{T}()$ to tackle with some records $R$ and attributes $S$ on a data table $D$ in the data lake. We show that this framework subsumes a number of commonly occurred tasks on structured data and can be easily extended to new and complex tasks even on unstructured data. (in Section~\ref{sec:problem})

2) \textbf{We abstract the general procedures to solve different tasks using LLMs.} We observe that, the key to solving a data manipulation task is to find a proper prompt to inspire the LLMs to produce accurate results~\cite{wang2022promptem}. However, due to the complexity of our tasks, it is difficult and ineffective to directly ask the LLMs for final results by a singleton prompt~\cite{narayan2022can}. As a result, we decompose a data manipulation task (that could be described by our unified framework) into several consistent steps such that each step is a simple, direct and easy job for LLMs. 

As illustrated in \autoref{fig:overview}, our solution contains three main steps. The first step automatically extracts relevant context information from data table to serve as demonstrations or background knowledge to solve the task. The second step transforms the context information from tabular form to logic text so the LLMs can more easily capture its semantics. Finally, the third step applies prompt engineering to construct the target prompt to obtain the final results.
In such a way, we attain generality across different tasks and effectiveness by LLMs. (in Section~\ref{sec:archi})

3) \textbf{We design effective (templates of) prompts for each main step in our solution.}
For each main step, we abstract the knowledge that needs to be acquired from LLMs and design a general template of prompt to automatically extract such knowledge. In such a way, LLMs could do well in each step and improve the quality of the final results. (in Section~\ref{sec:archi})

4) \textbf{We conduct extensive experiments to evaluate the performance of our solution \sys.} The evaluation results on lots of benchmarks exhibit that \sys attains state-of-the-art results on a variety of data manipulation tasks, including data imputation, data transformation, error detection and entity resolution. Meanwhile, the effectiveness of each main step in \sys is also verified by ablation study. (in Section~\ref{sec:exp})

\eat{
5) \textbf{We point out a number of future research directions.} 
Our work takes the first step on applying LLMs to design general solutions for data manipulation tasks on data lakes. There still reserves much room for improvement upon our work, including the extension to more types of data (e.g., domain-specific data, unstructured or semi-structured data), the new paradigm of large models tailored to database tasks and new methods with better efficiency and more friendly to system deployment. (in Section~\ref{sec:discussion})
}

\section{Background and Motivation}
\label{sec:back}

\sstitle{Large Language Models and Prompts.}
LLMs could be regarded as a foundation model applicable to numerous tasks, particularly for tasks requiring to interpret the semantics of data. Instead of fine-tuning the model to fit each task, we can simply apply \emph{prompts} to guide LLMs to solve each task. Specifically, a prompt is an intuitive interface written in natural text to interact with LLMs. It can take various forms (e.g., phrases or complex sentences) to guide or ask the LLMs to extract their knowledge to perform lots of different jobs, such as code generation, question answering, creative writing, etc. For example, a simple prompt, such as ``Translate English to French: hello =$>$'', could directly do the language translation.

The performance of the LLMs is very sensitive to the prompt~\cite{brown2020language}. To obtain high quality results, we often design prompts with context information to provide more instructions to LLMs. The context information could be a few input/output examples or other information relevant to the task. When combined with the task description, LLMs have more background to extract more accurate knowledge to answer the questions. For example, for the prompt ``Fill in the value like Genre: Folk; Artist: Bob Dylan. Genre: Jazz; Artist: ?'', the LLM would imitate the example to find a jazz artist, e.g., ``Bill Evans'', as a result. 

\sstitle{LLMs for Data Tasks.}
Some very recent works~\cite{RetClean,brunner2020entity,li2020deep,mei2021capturing,chensymphony,narayan2022can,wang2022promptem,Trummer2022b,trummer2022dbbert} observe the potential benefits of bringing LLMs into some data manipulation tasks. For example, we could ask LLMs to automatically judge whether a value is valid for an attribute using its intrinsic knowledge instead of designing numerous error detection rules for each domain. Prior works~\cite{herzig2020tapas,liu2021tapex,peeters2021dual,Trummer2022b} have verified the effectiveness of applying LLMs to answer questions on data tables. Later, LLMs have been applied on data pre-processing tasks~\cite{li2020deep}, binary classification on tables~\cite{TabLLM2022}, data cleaning and integration tasks\cite{RetClean, narayan2022can}. These works provide enough evidence to exhibit that the LLM-based methods could attain very promising, sometimes state-of-the-art, performance on these tasks.

However, even avoiding human efforts on providing domain knowledge, current LLM-based methods are not generally applicable to data manipulation tasks. The procedures, along with the prompts, of these methods are all specifically designed for each task, which require users to manually extract customized context information to guide the LLMs. The benefits of the LLMs and the shortcomings of existing methods motivate us to ask the following question:

\emph{Could we find a unified solution with LLMs such that it is both general and automatic to different data manipulation tasks on data lakes requiring no manual efforts?}

\section{Problem Definition}
\label{sec:problem}

In this section, we propose a unified framework to formalize the data manipulation tasks we target to solve on data lakes. Let $\mathcal{D} = \{D_1, D_2, \dots, D_l\}$ be a data lake. In this paper, we assume that each element $D_i \in \mathcal{D}$ is a relational data table containing a number of records (tuples). We denote the schema, as well as the set of attributes, of table $D_i$ as $S_i$. Unlike with the relational database, the join relations are not specified for tables in the data lake $\mathcal{D}$. For any record $r$ and attribute $s$, we denote the value of $r$ on $s$ as $r[s]$.

Let $T$ represent a data manipulation (e.g., data cleaning or integration) task on $\mathcal{D}$. We assume that the task description and the parameters of the task (e.g., a question on the table) are all encoded in $T$. We could formalize a number of different tasks in a unified manner as follows:

\textbf{Input}: a data lake $\mathcal{D}$, a subset of records $R \subseteq D_i$ extracted from a table $D_i \in \mathcal{D}$, a subset of attributes $S \subseteq S_i$ 
under the schema $S_i$ and a target task $T$;

\textbf{Output}: we have a function $F_{T}$ related to the task $T$ that produces a value $Y = F_{T}(R, S, \mathcal{D})$.

For each different data manipulation task $T$, the function $F_{T}$ is defined in different ways according to the applications. We list a number of example tasks that are commonly used in real-world applications and can be subsumed by our framework as follows:

\squishlist
\item{\bf Data Imputation}: This task is to repair dirty data and impute the missing value within a record. $S$ contains an attribute in $S_i$ and $R$ contains a singleton record in $D_i$ having a missing value on attribute $S$, $F_{T}(R, S, \mathcal{D})$ outputs the desired missing value of $R[S]$. 

\item{\bf Data Transformation}: This task is the process of converting data from one format to another required format within a record. $S$ contains an attribute in $S_i$ and $R$ contains a singleton record in $D_i$, $F_{T}(R, S, \mathcal{D})$ transforms the original value $R[S]$ to another new value $R'[S]$ by user specified rules.

\item{\bf Error Detection}: This task is to detect attribute error within a record in a data cleaning system. $S$ contains an attribute in $S_i$ and $R$ contains a singleton record in $D_i$, $F_{T}(R, S, \mathcal{D})$ predicts whether the value $R[S]$ is normal or not.

\item{\bf Entity Resolution}: This task is the process of predicting whether two records are referencing the same real-world thing. $S$ contains a number of attributes describing the property of each record in $D_i$ and $R = \{r_1, r_2\}$ contains two different records in $D_i$, $F_{T}(R, S, \mathcal{D})$ outputs whether the two records $r_1$ and $r_2$ refer to the same real-world entity or not.




\squishend

Notably, in our proposed framework, we just consider the very fundamental form for the data manipulation tasks in data lakes. In the following content, we apply the above data manipulation tasks subsumed by our framework to showcase how to design a general solution using LLMs. Whereas, our framework could be easily extended with rich definitions to support new tasks on unstructured or semi-structured data. 

\begin{figure*}[tbp]
  \centering
  \includegraphics[width=\linewidth]{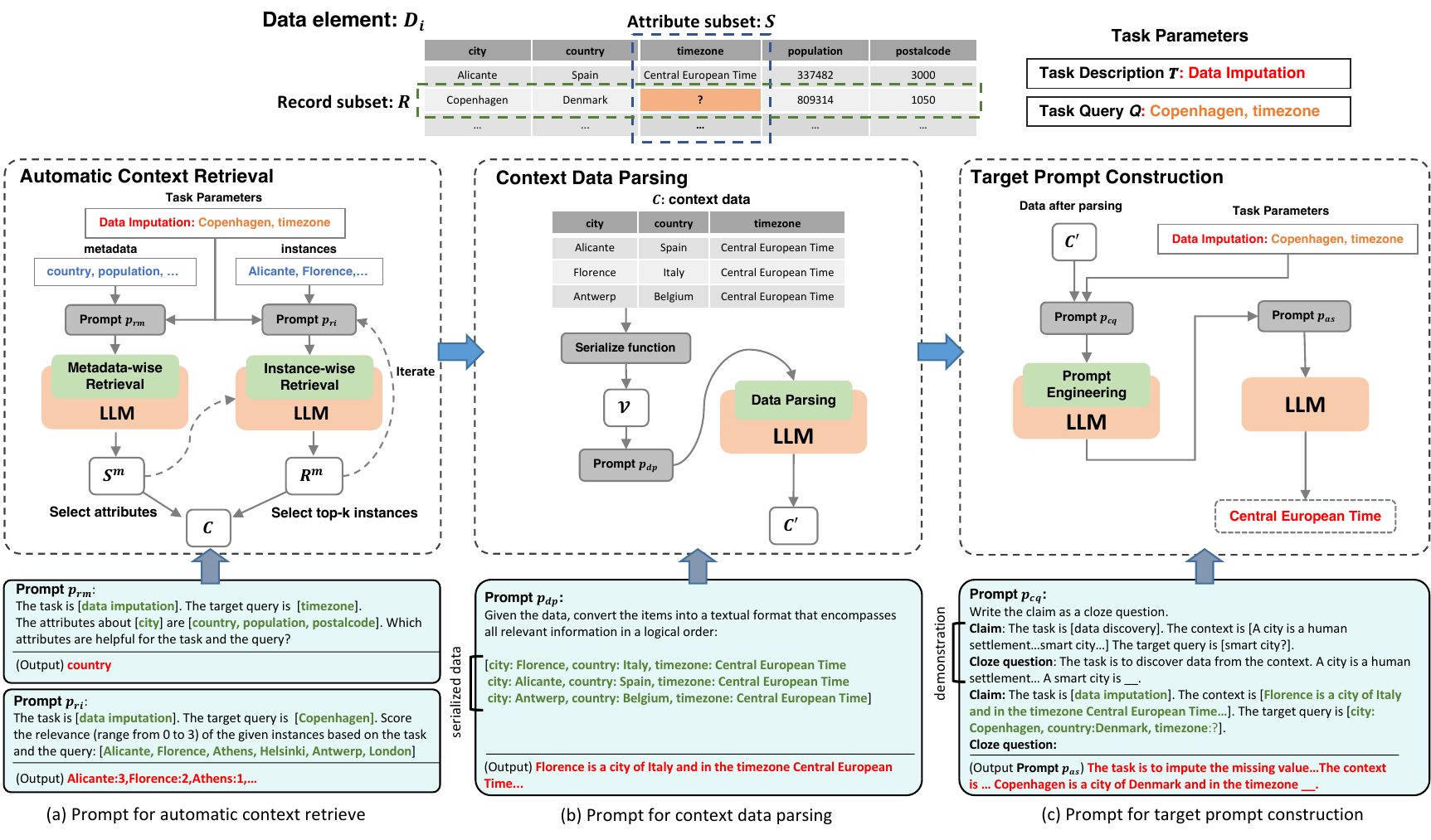}
  \caption{The pipeline of \sys by taking data imputation task as the example.}
  \label{fig:pipeline}
\end{figure*}

\section{Unified Framework for Data Manipulation}
\label{sec:archi}

To generally support common data manipulation tasks, we propose the \sys, a unified framework based on LLMs.
We first analyze this problem and outline our solution in Section~\ref{sec:archi-1}.
Then, Section~\ref{sec:archi-2} to Section~\ref{sec:archi-4} elaborate on the details of each key technique in our method. Finally, the generality of our method is discussed in Section~\ref{sec:archi-5}.

\subsection{Overview}
\label{sec:archi-1}

\sstitle{Problem Analysis.}
Recall that, we could consult the LLMs using a \emph{prompt} to acquire the desired knowledge. The data manipulation tasks could also be solved using proper prompts. As shown in the following example, we apply two simple prompts to resolve the data imputation and transformation tasks. Here each prompt is a textual template in natural language containing:
1) the task description (marked in red) so the LLMs could understand what to be done, e.g., filling in the missing value or transforming the data;
2) the context information (marked in blue) serving as the demonstrations or examples to guide the LLMs to perform the task, e.g., information of other relevant attributes or examples of the transformation;
and 3) the placeholder (marked in orange) to provide the input of the task. 

\vspace{0.5em}
\noindent\fbox{\parbox{0.98\linewidth}{
\textbf{Prompt A}:\\
\small{{\color{red} Data Imputation:} {\color{blue} city:Copenhagen,country:Denmark,} {\color{orange} timezone:?}} \\
\textbf{Prompt B}:\\
\small{{\color{red} Data Transformation:} {\color{blue} 20210315 to Mar 15 2021,} {\color{orange} 20201103 to ?}}
}}
\vspace{0.5em}

It has been shown that the quality of the LLM results is very sensitive to the format of the prompt~\cite{brown2020language}.
Simple and straightforward prompts (such as the above two examples) would often have mediocre performance on data manipulation tasks. This is because LLMs are often trained on the corpus to reason about direct and simple events, e.g. ``$x$ is contained in $y$'' or ``$y$ is the same as $z$'', but may fail on complex multi-hop reasoning problems that they have no explicit evidence~\cite{creswell2022selection}, e.g. ``Is $x$ contained in $z$?''. The process of solving a data manipulation task is too complicated for LLMs. It requires LLMs to interpret the task description, to extract (and possibly transform) the context information and to fill in a proper value in the placeholder at the same time.

Some literature works~\cite{chensymphony,wang2022promptem} have devoted the efforts on designing and tuning prompts for some specific data manipulation tasks. Whereas, this is very expensive and not extensible to numerous customized tasks occurring every day. What we actually need is a general solution that is applicable to produce effective prompts for each different data manipulation task.

\sstitle{Our Main Idea and Solution.}
To attain this goal, we decompose a data manipulation task (that could be described in Section~\ref{sec:problem}) into several consistent steps. The pipeline and architecture of \sys are illustrated in \autoref{fig:pipeline}. Each step is a simple and direct easy task to reason about for LLMs using its evidence and logic. Based on this, for each step, we abstract the knowledge that needs to be acquired from LLMs and design a general template of prompt to extract such knowledge. \sys takes data from a data lake as input and performs a data manipulation task in an iterative and interactive fashion. Generally, \sys proceeds a data manipulation task in three main steps: 

1) {\bf Context Retrieval}: 
Given the task $T$, the records $R \subseteq D_i$ and the attributes $S \subseteq S_i$ on a data lake $\mathcal{D}$, at the very beginning, we need to extract the relevant contextual information $\mathcal{C}$ from $\mathcal{D}$ to resolve $T$. $\mathcal{C}$ may contain additional information from other records and attributes to guide the LLMs to capture the semantics for task $T$.
We design two templates of prompts to automatically retrieve context information from $\mathcal{D}$ using LLMs. The first prompt $p_{rm}$ aims at retrieving meta-wise information, e.g., a number of attributes that may provide useful knowledge to task $T$ on attributes in $S$. Based on its results, the second prompt $p_{ri}$ identifies the most helpful records related to $R$ to resolve the task in a more fine-grained manner. After that, we obtain the context information $\mathcal{C}$ in a tabular form.

2) {\bf Context Parsing}:  
The raw context information in $\mathcal{C}$ in tabular form is often not friendly to be understood by the LLMs (as LLMs are mainly trained to interpret the natural language text). Therefore, the next step is to transform the original context $\mathcal{C}$ into another form that is more easily to be interpreted by the LLMs. Similar to previous works~\cite{narayan2022can}, we first apply a serialize function to transform $\mathcal{C}$ into a regular text $\mathcal{V}$ with pairs of attributes and values, e.g., ``city:Florence, country:Italy''. Then, we design a prompt template $p_{dp}$ to further convert the text $\mathcal{V}$ into the natural text $\mathcal{C}'$ reflecting the logic relations among different attributes, e.g. ``Florence is a city of Italy''. $\mathcal{C}'$ is smoother and closer to the natural language, so the LLMs could find more relevant information in its corpus for downstream procedures.

3) {\bf Target Prompt Construction}: 
Finally, we combine the serialized text $\mathcal{C}'$, the description of the task $T$ and the task input $R$ and $S$ together to consult the LLMs to 
obtain the final result $Y$. We utilize prompt engineering to automatically elicit prompts for any data manipulation task. Specifically, all data manipulation tasks described by our unified framework in Section~\ref{sec:problem} could be equivalently transformed into a cloze question. Cloze question is friendly to LLMs as it is written in natural language with placeholders. To automatically generate a proper cloze question, we design a template of prompt $p_{cq}$ that provides the LLMs a small set of demonstrations, where each one is a pair of a data manipulation task and its corresponding cloze question. These demonstrations include both task-specific and task-agnostic examples, where LLMs could learn to identify the most suitable template for any task to output a cloze question $p_{as}$. The target prompt $p_{as}$ is fed into LLMs to obtain the final result $Y$.

\eat{
In the following content, we describe the details of each component in our solution. For ease of presentation, we use the data imputation task as an example to introduce the technical details. We discuss how our solution generalizes to other data manipulation tasks in Section~\ref{sec:archi-5}.
}

\subsection{Automatic Context Retrieval}
\label{sec:archi-2}

To capture data knowledge from data lakes in a more interpretable and modular way, we augment LLMs with an automatic context retrieval component.
We require that the context retrieval component could identify useful information while filtering irrelevant data to facilitate the LLMs. 
Previous works~\cite{DataWig,narayan2022can} ask users to specify the instances (records) and attributes relevant to the task or learn to identify useful records based on the similarity of attribute values~\cite{mei2021capturing,RetClean}. Unlike with them, we design a purely automatic strategy for extracting helpful attributes and records with the aid of the LLMs. We first apply meta-wise retrieval to find relevant attributes from the holistic view. Then we apply instance-wise retrieval to extract useful records in a more detailed manner.

\sstitle{Meta-wise Retrieval.}
In the first step, we provide a number of candidate attributes $S'$ and ask LLMs to select valuable ones for our task $T$ and target attribute $S$. LLMs could apply the inherent knowledge to measure the relationships between $S'$ and $S$ and reserve only promising attributes. This step could filter irrelevant information in the data lake $\mathcal{D}$ in a coarse-grained manner. The results would contain helpful meta-wise information describing the high level domain knowledge of these attributes. Specifically, we construct the prompt $p_{rm}$ using the following template:

\vspace{1.2em}
\noindent\fbox{\parbox{0.98\linewidth}{
\textbf{Prompt $p_{rm}$}: \\
\textsl{The task is $[T]$. The target query is $[Q]$. The candidate attributes are $[s_1,s_2,...,s_n]$. Which attributes are helpful for the task and the query?}
}}
\vspace{1.2em}

Here $T$ is the task description such as ``data imputation''. The query $Q$ combines our inputs on the target record $R$ and attribute $S$ for task $T$. It has different forms in different tasks. In data imputation, we represent it as ``the primary key of record $R$, the attribute $S$'', e.g., ``Copenhagen, timezone'', to indicate that we want to fill the value of $S$ for record $R$. The set of candidate attributes $S' = S_i - S = \{s_1, s_2, \dots, s_n\}$ contains all remaining attributes in $S_i$ of table $D_i$. For other tasks, the forms of the query $Q$ and set $S'$ are different. We reserve the details in  Section~\ref{sec:archi-5}. In this paper, we do not apply cross-table attributes in table $D_{j \neq i} \in \mathcal{D}$ as the experimental results on $D_i$ (shown in Section~\ref{sec:exp}) are competitive enough. 

We denote $S^t$ as the task-relevant attributes returned by the LLMs. In our example in \autoref{fig:pipeline} (left), when given the task description ``data imputation'' and the target query (attribute) ``timezone'', the LLMs select the attribute ``country'' towards inferring about the missing value.

\sstitle{Instance-wise Retrieval.}
Next, we perform fine-grained filtering on records to identify relevant ones w.r.t.~target records in $R$. 
We first shrink the data in $D_i - R = \{r_1, r_2, \dots, r_m\}$ to provide a set of candidate records $R'$ by random-sampling. After that, we use the LLMs to examine the relevance between $R'$ and $R$. The \emph{relevance} can be interpreted in different ways for different tasks. For example, for data imputation, we hope to find records similar to target records $R$ and attributes $S$ to find the missing value. For the error detection task, we may want to obtain records reflecting the distribution of the domain value to identify whether the target value $R[S]$ is abnormal. 
We still drive LLMs to consult the semantic knowledge to measure the relevance scores of the records for the target task by a prompt $p_{ri}$ using the following template:

\vspace{1.2em}
\noindent\fbox{\parbox{0.98\linewidth}{
\textbf{Prompt $p_{ri}$}: \\
\textsl{The task is $[T]$. The target query is $[Q]$. To score the relevance (range from 0 to 3) of given instances based on the task and the query: $\{r_1[S^{t}],r_2[S^{t}],...,r_m[S^{t}]\}$}
}}
\vspace{1.2em}

Here for each $r_j \in R'$, we only reserve the task-relevant attributes in $S^t$ as the other ones are identified to be not helpful to our task. After examining all records in $R'$ (or touching a time limit), we order all instances according to the relevance score and select the top-$k$ instances $R^t$ as the task-relevant context $\mathcal{C}$ for the downstream procedures. 
\subsection{Context Data Parsing}
\label{sec:archi-3}

The context information $\mathcal{C}$, represented in a tabular form, is not friendly for the LLMs to interpret its underlying semantics, since the LLMs are often trained on large corpus of text. To resolve this problem, we consider how to transform $\mathcal{C}$ into a more effective format for LLMs. As all records $R^t$ in $\mathcal{C}$ are all organized in a regular structure under the schema $S^t$, we could easily serialize $\mathcal{C}$ into a textual string. Specifically, let $\{(s, r[s]) | r \in R^t, s \in S^t\}$ denote the set of all pairs of each attribute $s$ and its value in record $r$. The information of $\mathcal{C}$ is losslessly encoded. Our $serialize()$ function directly concatenates all pairs to produce a text $\mathcal{V}$. 

Previous works~\cite{narayan2022can} directly feed the text $\mathcal{V}$ into LLMs to serve as the contextual information for our task. Whereas, we further try to integrate the pairs in $\mathcal{V}$ into a logic text $\mathcal{C}'$ reflecting the relations among different attributes. For example, in Figure~\ref{fig:pipeline} (middle), the text ``country:Italy, timezone: Central European Time'' is converted into ``The country Italy is in the Central European Time timezone''. Obviously, the former one rarely occurs in any article except some tables while the latter one may frequently occur in some scientific articles in the training corpus. Therefore, providing $\mathcal{C}'$ rather than $\mathcal{V}$ to LLMs could improve its probability of hitting relevant texts in inference and produce more accurate results. 

Notably, converting the text $\mathcal{V}$ to $\mathcal{C}'$ is an easy job for LLMs. The logic relations among different attributes are often common and fixed, e.g. ``a city is in a country in a timezone'', so the LLMs could directly capture such knowledge. In our solution, we apply the following data parsing template prompt $p_{dp}$ to perform this job. The generated context representation $\mathcal{C}'$ is applied in the subsequent procedures.

\vspace{1.2em}
\noindent\fbox{\parbox{0.98\linewidth}{
\textbf{Prompt $p_{dp}$}:
\textsl{Given the data, convert the items into a textual format that encompasses all relevant information in a logical order: $[\mathcal{V}]$}
}}
\vspace{1.2em}

\subsection{Effective Target Prompt Construction}
\label{sec:archi-4}

To apply LLMs for our task, the ultimate (as well as the most important) step is to find an effective prompt to organize the task description $T$, the context information $\mathcal{C}'$ in logic text and the query $Q$ encoding the input records $R$ and attributes $S$ (defined in Section~\ref{sec:archi-2}) together. Moreover, we hope that the method to find the prompt is generally applicable to different tasks to avoid exhaustive tuning efforts. 

We observe that, all of our tasks described by the claims (containing $T$, $C'$, $R$ and $S$ as stated above) could be equivalently summarized as a cloze question. Specifically, cloze question asks the model to fill in the remaining text (``Australia and Switzerland won \_\_ gold medals in total.''). Cloze question is friendly to LLMs as it is written in natural language with placeholders. For example, the data imputation task is to fill the missing value of $R[S]$ and the error detection task is to fill a normal or abnormal answer for the value $R[S]$. Therefore, our problem is how to automatically organize our claims for different tasks into a proper cloze question. 

Certainly, it requires \sys to capture the semantics of each element in our claims, e.g., which element should be placed in front of others, and organize them in a smooth natural text. In similar to context data parsing, this job is suitable to be done by the LLMs themselves. We apply the following prompt $p_{cq}$ to do this transformation:


\vspace{1.2em}
\noindent\fbox{\parbox{0.98\linewidth}{
\textbf{Prompt $p_{cq}$}:\\
\textsl{
Write the claim as a cloze question. \\
\textbf{Claim}: The task is data imputation. The context is...\\
\textbf{Cloze question}: ... China's population is \_\_. \\
\textbf{Claim}: The task is data transformation. The context is...\\
\textbf{Cloze question}: ... The roman numeral III can be transformed to normal number \_\_. \\
......\\
\textbf{Claim}: The task is $[T]$. The context is $[\mathcal{C}']$. The target query is $[Q]$. \\
\textbf{Cloze question}:
}}}
\vspace{1.2em}

Motivated by the discrete prompt search methods~\cite{gao2021making,arora2022ask}, we provide a number of claims and their corresponding cloze questions as demonstration examples in $p_{cq}$. The pairs of claim and cloze question include:
1) examples pertaining to our commonly applied tasks (such as data imputation and error detection) that are verified to produce accurate results;
and 2) some task-agnostic transformation strategies which are verified to be generally applicable to different tasks.
LLMs could learn from these examples to identify the most suitable template to transform our claims on a task to a cloze question $p_{as}$. In such a way, we attain both high effectiveness (on common tasks) and cross-task generality (on new and unseen tasks) in generating prompts. We only need to maintain the demonstration examples according to the applications in periodical while avoiding specialized prompt design for each upcoming task.

Finally, we feed the prompt $p_{as}$ into LLMs to yield the final answer of our task. The experimental results in Section~\ref{sec:exp} indicate that the prompts generated by our method are very effective on a variety of data manipulation tasks.

\subsection{Generalization to More Tasks}
\label{sec:archi-5}

Our \sys framework could be easily generalized to other tasks listed in Section~\ref{sec:problem} by minor adaptions to the form of the query $Q$, which encodes the target records $R$ and attributes $S$, and the set $S'$ of candidate relevant attributes in prompt $p_{rm}$ (defined in Section~\ref{sec:archi-2}). On the other hand, our method can flexibly combine various modules for different tasks. The details are listed as follows. 

For the data transformation task, we directly set $Q = R[S]$ to give the attribute value to be transformed. For error detection, we represent it as ``$S$: $R[S]$?'' to indicate whether $R[S]$ is a valid value for $S$.
For entity resolution where $R = \{r_1, r_2\}$, we set $Q$ to be ``Entity A is $r_1$, Entity B is $r_2$'' to identify whether $r_1$ and $r_2$ refer to the same entity. 



 
For other tasks that could be subsumed by our framework defined in Section~\ref{sec:problem}, we could also adjust the parameters and module combination according to the semantics of the task.

 
\eat{
To understand the cross-task generalization ability of \sys, we extend our framework to more tasks. All tasks are built in the same pipeline. The difference is the parameter of the task that contains a task description $T$ and a query $Q$. We list the parameter forms for a number of example tasks that can be subsumed by our framework as follows:

\noindent\textbf{Data Transformation}: The query $Q$ parameterizes inputs on the target record $R$ and attribute $S$. We represent it as ``the attribute $S$ of record $R$ to be transformed'', e.g., ``20201103''.

\noindent\textbf{Error Detection}: The query $Q$ parameterizes inputs on the target record $R$ and attribute $S$. We represent it as ``the attribute $S$ of record $R$ to classify as an error'', e.g., ``CountyName: jefferson?''.

\noindent\textbf{Entity Resolution}: The query $Q$ parameterizes inputs on two records $R=\{r_1, r_2\}$. We represent it as ``records $r_1$ and $r_2$ to classify as the same entity'', e.g., ``Entity A is a song..., Entity B is a song...''.

\noindent\textbf{Table Question Answering}: The query $Q$ parameterizes inputs on a natural language question. We represent it as ``question to be executed against the records'', e.g., ``how many gold medals did Australia and Switzerland total?''.
}


\section{Experiment}
\label{sec:exp}

In this section, we conduct extensive experiments on different data manipulation tasks to evaluate the generalization and quality of our \sys.
We also perform an in-depth analysis of \sys on more data types and task forms.
And then, we provide several model variants to show the effectiveness of the proposed method. 

\subsection{Experimental Setup}
\label{sec:exp-1}

\sstitle{Implement Details.}
We implement our \sys using the GPT-3-175B parameter model~\cite{brown2020language}(text-davinci-003) in the OpenAI API~\cite{OpenAI} as the LLM without fine-tuning. In addition, we give fine-tuning results for open-source LLMs GPT-J-6B~\cite{gptj} and LLaMA2-7B~\cite{llama2}. In our method, in the default setting, we apply the automatic context retrieval method proposed in Section~\ref{sec:archi-2}. In detail, we extract one attribute from the candidate set in the metadata-wise retrieval (see prompt $p_{rm}$ in Section~\ref{sec:archi-2}) and top-3 records from 50 records randomly sampled in the dataset in the instance-wise retrieval (see prompt $p_{ri}$ in Section~\ref{sec:archi-2}). 

\sstitle{Evaluation Tasks and Datasets.}
We evaluate \sys on a number of different data manipulation tasks including data imputation, data transformation, error detection and entity resolution. 
We evaluate the performance of our method on different benchmark datasets. 
For data imputation, we choose two challenging benchmark datasets, namely Restaurants and Buy, from~\cite{mei2021capturing}. For Restaurants dataset, the target attribute to be impute is ``city''. For Buy dataset, the target attribute to be impute is ``manufacturer''. We manually mask the values in the target attributes. Ground truth information is available for the missing values.  
For data transformation, we follow the TDE benchmark in~\cite{tde2018} and choose two datasets, namely StackOverflow and Bing-QueryLogs. This benchmark covers diverse types of transformation tasks (e.g., ip, address, phone, etc).
For error detection task, we choose the benchmark Hospital and Adult datasets widely used in data cleaning papers~\cite{rekatsinas2017holoclean,heidari2019holodetect}. Errors amount to $~$5\% of the total data. Ground truth information is available for all cells.  
For entity resolution, we follow the standard Magellan benchmark in~\cite{konda2016magellan} and choose four datasets across different domains. Each dataset consists of candidate pairs from two structured tables of entity records of the same schema. The ground truth labels (positive or negative) are available for the entity pairs.

\eat{
\begin{table}[!tbp]
\centering
\caption{Basic domain information of evaluation dataset on 6 data manipulation tasks.}
\label{tab:dataset}
\begin{tabular}{c|c|c}
\textbf{Task}                        & \textbf{Dataset} & \textbf{Domain} \\
\thickhline
\multirow{2}{*}{Data Imputation}     & Restaurant       & restaurant      \\
                                     & Buy              & product         \\ \hline
\multirow{2}{*}{Data Transformation} & StackOverflow    & code snippet   \\
                                     & Bing-QueryLogs   & unit conversion \\ \hline
\multirow{2}{*}{Error Detection}     & Hospital         & medicine        \\
                                     & Adult            & person          \\ \hline
\multirow{4}{*}{Entity Resolution}   & Beer             & product         \\
                                     & Amazon-Google    & software        \\
                                     & iTunes-Amazon    & music           \\
                                     & Walmart-Amazon   & electronics     \\ \hline
Join Discovery                       & NextiaJD         & N$/$A  \\ \hline
Information Extraction               & NBA player       & sport          
\end{tabular}
\end{table}
}

\sstitle{Baseline Approaches.}
We compare \sys with a variety of state-of-the-art (SOTA) methods on data manipulation tasks. The \textbf{FM}~\cite{narayan2022can} method is shown to attain SOTA performance on multiple data manipulation tasks with simple prompt learning on LLMs. We reproduce FM following the original paper and its open-source code. In its default setting, the context information and the target prompt are manually selected by guiding rules and it only applies serialization in context data parsing. We evaluate FM on data imputation, data transformation, error detection and entity resolution tasks. We also reproduce a random-sample version of FM, where the context information of records is randomly selected from the table. 

For data imputation task, we select several methods following different technical routines, including a statistics-based method \textbf{HoloClean}~\cite{rekatsinas2017holoclean,wu2020attention}, a clustering-based method \textbf{CMI}~\cite{CMI2008} and a deep learning based method \textbf{IMP}~\cite{mei2021capturing}. For data transformation task, we select a search-based method \textbf{TDE}~\cite{tde2018}. For error detection task, we select two machine learning based methods \textbf{HoloClean}~\cite{rekatsinas2017holoclean} and \textbf{HoloDetect}~\cite{heidari2019holodetect}. For entity resolution task, we use a deep learning method \textbf{Ditto}~\cite{li2020deep}. 

\sstitle{Evaluation Metrics.} Following previous works, we employ widely-used metrics, \textit{accuracy}, \textit{precision}, \textit{recall} and \textit{F1-score} to evaluate the effectiveness of these methods. For data imputation and data transformation, we use accuracy to denote the fraction of correct repairs over the total number of repairs performed for cells in the labeled data. For error detection and entity resolution, we use F1-score based on precision and recall.

\sstitle{Evaluation Goals.}
The experimental results mainly answer the following questions: 
\squishlist
\item{What is the performance of our \sys solution on different data manipulation tasks? (in Section~\ref{sec:exp-2})}
\item{What is the contribution of each component in our \sys solution? (in Section~\ref{sec:exp-3})}
\squishend

\subsection{Performance Evaluation}
\label{sec:exp-2}

\sstitle{Data Imputation.} As shown in \autoref{tab:di}, we conduct experiments to compare \sys with other methods. Here we also compare the performance of \sys and FM with another setting, where the context information of records is randomly selected from the table. We find that:

\begin{table}[!tbp]
\centering
\caption{Accuracy on data imputation task with SOTA.}
\label{tab:di}
\begin{tabular}{c|cc}
\multirow{2}{*}{\textbf{Method}} & \multicolumn{2}{c}{\textbf{Data Imputation Accuracy (\%)}}            \\
                  & \multicolumn{1}{c|}{\textbf{Restaurant}} & \textbf{Buy} \\ 
\thickhline
HoloClean        & \multicolumn{1}{c|}{33.1}                & 16.2         \\
CMI              & \multicolumn{1}{c|}{56.0}                & 65.3         \\
IMP              & \multicolumn{1}{c|}{77.2}                & 96.5         \\
FM (random)      & \multicolumn{1}{c|}{81.4}                & 86.2         \\
FM (manual)              & \multicolumn{1}{c|}{88.4}                & 98.5\\
\sys (random)  & \multicolumn{1}{c|}{87.2}                & 92.3         \\
\sys           & \multicolumn{1}{c|}{93.0}       & 98.5
\end{tabular}
\vspace{-1em}
\end{table}

\begin{table}[!tbp]
\centering
\caption{Accuracy on data transformation task with SOTA. }
\label{tab:dt}
\begin{tabular}{c|cc}
\multirow{2}{*}{\textbf{Method}} & \multicolumn{2}{c}{\textbf{Data Transformation Accuracy (\%)}}                      \\
                                 & \multicolumn{1}{c|}{\textbf{StackOverflow}} & \textbf{Bing-QueryLogs} \\
\thickhline
TDE                              & \multicolumn{1}{c|}{63.0}                   & 32.0                    \\
FM                               & \multicolumn{1}{c|}{65.3}                   & 54.0                    \\
\sys                              & \multicolumn{1}{c|}{67.4}          & 56.0                       
\end{tabular}
\vspace{-1em}
\end{table}

\begin{table}[!tbp]
\centering
\caption{F1-score on error detection task with SOTA. }
\label{tab:ed}
\begin{tabular}{c|cc}
\multirow{2}{*}{\textbf{Method}} & \multicolumn{2}{c}{\textbf{Error Detection F1-Score (\%)}} \\
                                 & \multicolumn{1}{c|}{\textbf{Hospital}}       & \textbf{Adult} \\ 
\thickhline
HoloClean                        & \multicolumn{1}{c|}{51.4}                    & 54.5        \\
HoloDetect                       & \multicolumn{1}{c|}{94.4}                    & 99.1        \\
FM                               & \multicolumn{1}{c|}{97.1}                    & 99.1        \\
\sys                             & \multicolumn{1}{c|}{99.8}                    & 99.7              
\end{tabular}
\vspace{-1em}
\end{table}

1) Overall, \sys attains significantly higher accuracy than the SOTA results. Although FM applies costly manual selection of context information, the accuracy of \sys is still $4.6\%$ higher than FM on Restaurant dataset and comparable on Buy dataset. This verifies the effectiveness of our \sys solution, especially the automatic retrieval of the context information.

2) For the random-setting with the same context information selected randomly from the table, \sys still outperforms FM by $5.8\%$ on the Restaurant dataset and $6.1\%$ on the Buy dataset. This is because \sys applies logic transformation of the context information, rather than only simple serialization employed in FM. Meanwhile, \sys searches for the most effective target prompt by utilizing the knowledge in the LLM, rather than simple construction by users. This verifies the success on our design choices of the context data parsing and target prompt construction.

\sstitle{Data Transformation.}
\sys also achieves the most promising results on the data transformation task. As shown in~\autoref{tab:dt}, \sys outperforms the search-based method TDE and the LLM-based FM. In comparison to FM (the current SOTA results), \sys yields nearly $2\%$ gain on the two datasets in terms of the accuracy. The reasons are analyzed in the above experiment. This also verifies the advantages of LLM-based methods over other approaches. 

\sstitle{Error Detection.}
We report the F1-score obtained by \sys and competing approaches.
\sys also achieves a similar behavior on the error detection task. As shown in~\autoref{tab:ed}, \sys outperforms the baseline methods HoloClean, HoloDetect and FM by up to $2.7\%$ in terms of the F1-score in the Hospital dataset. For the Adult dataset, \sys achieves a high F1-score of 99.7\%, as it uses the information on data source. It proves that our method is useful in interpreting the domain knowledge to detect errors.

\sstitle{Entity Resolution.}
As shown in \autoref{tab:em}, \sys is also effective on the entity resolution task. In comparison to FM with random context information, \sys always attains higher (or at least comparable) accuracy. In comparison to Magellan and Ditto which fine-tunes the model with large amounts of task-specific labeled data, \sys still achieves comparable or better results in most cases. Sometimes, the accuracy of \sys is lower than Ditto and FM with manually selected context. This is because these datasets contain very specific domain words that do not commonly occur in the corpus. As a result, the LLMs have little knowledge on their semantics and may make errors in inference. A similar phenomenon is also observed in~\cite{narayan2022can}. To avoid this, Ditto utilizes domain data to fine-tune the model and FM manually selects instances to learn the domain knowledge.

\begin{table}[!tbp]
\setlength\tabcolsep{4pt}
\centering
\caption{F1-score on entity resolution task with SOTA. }
\label{tab:em}
\begin{tabular}{c|cccc}
\multirow{2}{*}{\textbf{Method}} & \multicolumn{4}{c}{\textbf{Entity Resolution F1-Score (\%)}} \\
 & \multicolumn{1}{c|}{\textbf{Beer}} & \multicolumn{1}{c|}{\textbf{\begin{tabular}[c]{@{}c@{}}Amazon-\\ Google\end{tabular}}} & \multicolumn{1}{c|}{\textbf{\begin{tabular}[c]{@{}c@{}}iTunes-\\ Amazon\end{tabular}}} & \textbf{\begin{tabular}[c]{@{}c@{}}Walmart-\\ Amazon\end{tabular}} \\ 
\thickhline
Magellan   & \multicolumn{1}{c|}{78.8} & \multicolumn{1}{c|}{49.1} & \multicolumn{1}{c|}{91.2}& 71.9 \\
Ditto      & \multicolumn{1}{c|}{94.4} & \multicolumn{1}{c|}{75.6} & \multicolumn{1}{c|}{97.1}& 86.8 \\
FM(random) & \multicolumn{1}{c|}{92.3} & \multicolumn{1}{c|}{60.7} & \multicolumn{1}{c|}{96.3}& 73.8 \\
FM(manual) & \multicolumn{1}{c|}{100}  & \multicolumn{1}{c|}{63.5} & \multicolumn{1}{c|}{98.2}& 87.0 \\
\sys      & \multicolumn{1}{c|}{96.3} & \multicolumn{1}{c|}{64.3} & \multicolumn{1}{c|}{96.3}& 88.2      
\end{tabular}
\vspace{-0.5em}
\end{table}

\begin{table}[!hbp]
\centering
\vspace{-0.5em}
\caption{Fine-tuning experiments: F1-score of \sys on entity resolution task (Walmart-Amazon dataset).}
\label{tab:ft}
\begin{tabular}{c|cc}
\multirow{2}{*}{\textbf{LLM}} & \multicolumn{2}{c}{\textbf{F1-Score (\%)}}     \\
                             & \multicolumn{1}{c|}{\textbf{FM}} & \textbf{\sys} \\
\thickhline
GPT-J-6B   & \multicolumn{1}{c|}{17.6}        & 17.8   \\
GPT-J-6B (fine-tune)  & \multicolumn{1}{c|}{84.2}        & 86.6  \\
LLaMA2-7B   & \multicolumn{1}{c|}{NA}        & 40.6  \\
LLaMA2-7B (fine-tune)& \multicolumn{1}{c|}{NA}        & 89.4  \\
GPT-3-175B & \multicolumn{1}{c|}{87.0}        & 88.2 
\end{tabular}
\vspace{-0.5em}
\end{table}

For fairness, we also conduct a lightweight fine-tuning on our \sys as LLMs' model capacity scaling, specifically targeting the entity resolution task.
We conduct the lightweight fine-tuning experiments based on the HuggingFace\cite{wolf2020transformers} library. In this setting, we freeze most of the pre-trained parameters and augment the model with a small trainable head~\cite{ding2021openprompt}. For the fine-tuning process, we use the training split of the Walmart-Amazon dataset, which consists of 6144 tuples. During fine-tuning, we use an AdamW optimizer and a cosine annealing learning rate scheduler with the linearly warm-up step of 100, initial learning rate of 4e-5 and final learning rate of 1e-5. Our model is trained over 8 V100 GPUs for 30 epochs with a batch size of 16.
We also reproduced this experiment on FM under the same setting.

We scale LLMs parameter size from 175B down to 6B/7B. As shown in \autoref{tab:ft}, the fine-tuned 6B/7B LLM exhibits comparable performance to the much larger 175B model, suggesting \sys has the potential to scale to even smaller models with proper fine-tuning. Meanwhile, on the fine-tuned small model, \sys performs better than FM. This indicates that our \sys could also attain higher accuracy by fine-tuning. 

To showcase the performance of our \sys across different base models, we further evaluate \sys over five LLM variants: GPT-4-Turbo~\cite{OpenAI}, Claude2 (about 100B)~\cite{claude2}, LLaMA2 (7B and 70B)~\cite{llama2}, and Qwen-7B~\cite{Qwen} on a data imputation task. We observe consistent high performance with \sys across these different base LLMs, demonstrating its adaptability and robustness. As shown in \autoref{tab:LLMs}, even on 7B models, our \sys maintains impressive performance, and it can achieve better results on larger models at a higher cost.

In~\autoref{tab:token}, we compare the LLMs' tokens between the FM and our \sys. It is evident that our method incurs greater token consumption than the FM. However, our \sys automates the processes of context retrieval and target prompt construction, which significantly reduces the human labor. 

\begin{table}[!tb]
\centering
\caption{\sys results on data imputation with LLMs variants.}
\label{tab:LLMs}
\begin{tabular}{c|cc}
\multirow{2}{*}{Model} & \multicolumn{2}{c}{Data Imputation Acc (\%)} \\
                       & \multicolumn{1}{c|}{Restaurant}       & Buy       \\
\thickhline
GPT-3-175B             & \multicolumn{1}{c|}{93.0}             & 98.5      \\ 
GPT-4-Turbo            & \multicolumn{1}{c|}{96.5}             & 98.5      \\
Claude2                & \multicolumn{1}{c|}{89.5}             & 96.9      \\ 
LLaMA2-7B              & \multicolumn{1}{c|}{86.0}             & 95.4      \\ 
LLaMA2-70B             & \multicolumn{1}{c|}{88.4}             & 96.9      \\
Qwen-7B                & \multicolumn{1}{c|}{86.0}             & 93.8   
\end{tabular}
\vspace{-0.5em}
\end{table}

\begin{table}[!tb]
\centering
\caption{Token consumption (per-query) comparison with FM.}
\label{tab:token}
\begin{tabular}{c|cc}
\multirow{2}{*}{Method}              & \multicolumn{2}{c}{Token Consumption}  \\
                                     & \multicolumn{1}{c|}{Restaurant} & Buy  \\ 
\thickhline
FM                                   & \multicolumn{1}{c|}{174}        & 246  \\
\sys (w/o retrieval) & \multicolumn{1}{c|}{325}        & 384  \\
\sys                                & \multicolumn{1}{c|}{6860}       & 7323
\end{tabular}
\vspace{-1em}
\end{table}

\subsection{Impact of Model Components}\label{sec:exp-3}   

\begin{table}[!tb]
\setlength\tabcolsep{2pt}
\centering
\caption{Ablation study of \sys on data imputation task (Restaurant dataset).}
\label{tab:ablation1}
\scalebox{0.85}{
\begin{tabular}{c|c|c|c|c}
\begin{tabular}[c]{@{}c@{}}\textbf{Instance-wise}\\ \textbf{Retrieval}\end{tabular} & \begin{tabular}[c]{@{}c@{}}\textbf{Meta-wise}\\ \textbf{Retrieval}\end{tabular} & \begin{tabular}[c]{@{}c@{}}\textbf{Target Prompt}\\ \textbf{Construction}\end{tabular} & \begin{tabular}[c]{@{}c@{}}\textbf{Context Data}\\ \textbf{Parsing}\end{tabular} & \textbf{Acc (\%)}    \\ 
\thickhline
           &            &            &                      & 82.6 \\
\checkmark &            &            &                      & 84.9 \\
           & \checkmark &            &                      & 90.7 \\
\checkmark & \checkmark &            &                      & 90.7 \\
\checkmark & \checkmark & \checkmark &                      & 91.9 \\
\checkmark & \checkmark & \checkmark & \checkmark  & 93.0
\end{tabular}
}
\end{table}

\begin{table}[!tb]
\setlength\tabcolsep{2pt}
\centering
\caption{Ablation study of \sys on data imputation task (Buy dataset).}
\label{tab:ablation2}
\scalebox{0.85}{
\begin{tabular}{c|c|c|c|c}
\begin{tabular}[c]{@{}c@{}}\textbf{Instance-wise}\\ \textbf{Retrieval}\end{tabular} & \begin{tabular}[c]{@{}c@{}}\textbf{Meta-wise}\\ \textbf{Retrieval}\end{tabular} & \begin{tabular}[c]{@{}c@{}}\textbf{Target Prompt}\\ \textbf{Construction}\end{tabular} & \begin{tabular}[c]{@{}c@{}}\textbf{Context Data}\\ \textbf{Parsing}\end{tabular} & \textbf{Acc (\%)}    \\ 
\thickhline
           &            &            &                      & 90.8 \\
\checkmark &            &            &                      & 92.3 \\
           & \checkmark &            &                      & 90.8 \\
\checkmark & \checkmark &            &                      & 92.3 \\
\checkmark & \checkmark & \checkmark &                      & 96.9 \\
\checkmark & \checkmark & \checkmark & \checkmark  & 98.5
\end{tabular}
}
\end{table}

\begin{table}[!tb]
\centering
\caption{Ablation study of \sys on data transformation task.}
\label{tab:ablation3}
\scalebox{0.85}{
\begin{tabular}{cc|cc}
\multirow{2}{*}{\begin{tabular}[c]{@{}c@{}}\textbf{Target Prompt}\\ \textbf{Construction}\end{tabular}} & \multirow{2}{*}{\begin{tabular}[c]{@{}c@{}}\textbf{Context Data}\\ \textbf{Parsing}\end{tabular}} & \multicolumn{2}{c}{\textbf{Data Transformation Acc} (\%)}    \\
 & & \multicolumn{1}{c|}{\begin{tabular}[c]{@{}c@{}}\textbf{Stack}\\ \textbf{Overflow}\end{tabular}} & \begin{tabular}[c]{@{}c@{}}\textbf{Bing}\\ \textbf{QueryLogs}\end{tabular} \\ 
\thickhline
 &   & \multicolumn{1}{c|}{63.3} & 52.0 \\
 & \checkmark & \multicolumn{1}{c|}{65.3} & 52.0 \\
\checkmark    &   & \multicolumn{1}{c|}{65.3} & 54.0 \\
\checkmark    & \checkmark   & \multicolumn{1}{c|}{67.4} & 56.0
\end{tabular}
}
\vspace{-0.5em}
\end{table}

For ablation study, we analyze the effectiveness of each component in \sys. Specifically, we disable one or more components in \sys and compare the performance of \sys with its variants. The results are shown in \autoref{tab:ablation1}, \autoref{tab:ablation2} and \autoref{tab:ablation3}. Our findings are described as follows:

\sstitle{Context Retrieval.}
When disable context retrieval component, \sys randomly samples the same number of attributes and/or records from the table as context information. At this time, the accuracy of \sys significantly decreases. We observe that, for the data imputation task, by simply using instance-wise and meta-wise retrieval, the accuracy of \sys could improve $2.3\%$ and $8.1\%$ on Restaurant dataset, respectively. Additionally, on the Buy dataset, using context retrieval also improves the accuracy by $1.5\%$. This is because our context retrieval leverages LLMs to capture relevant attributes and/or instances with semantic relationships, which provides richer background knowledge for the target prompt.
For the data transformation task, it does not require to extract context data and this step is not included in the ablation study.

\sstitle{Context Data Parsing.}
Without context data parsing, we only apply the serialization function to convent the tabular context information into a string. We observe that, for the data imputation task, data parsing also helps to improve the result accuracy, with a $1.1\%$ increase on the Restaurant dataset and a $1.6\%$ increase on the Buy dataset. Similarly, for the data transformation task, using data parsing improves the accuracy by $2\%$ on the StackOverflow dataset. This is because our data parsing bridges the gap between structured tabular data and natural language representation to make the context information more friendly to be interpreted by LLMs.

\sstitle{Target Prompt Construction.}
We also compare \sys with the simple prompt that directly combines the task description, the context information and the task inputs to obtain the final result. For the data imputation task, we find that our constructed target prompt using cloze questions improves accuracy by $1.2\%$ on the Restaurant dataset and by $4.6\%$ on the Buy dataset. Furthermore, for the data transformation task, the accuracy could also improve $2\%$ on both the StackOverflow and Bing-QueryLogs datasets by using this method. This verifies the effectiveness of our target prompt construction method. It learns from examples to identify the most suitable prompt, rather than the simple combination without semantic connections.

\subsection{Discussion}
\label{sec:exp-5}

The principal objective of our experimental study is to demonstrate the effectiveness of LLMs in enhancing data lakes and minimizing human effort. The challenge of dealing with massive amounts of data is a common issue faced by data lake systems. Data relationships take on various forms. For instance, in some cases, we may have shared values or keys; in others, the data may be complementary and thus have no value-overlap at all. To address this challenge, our method utilizes LLMs to understand data relationships, integrate heterogeneous data sources, and automatically identify the relational data for data tasks. Our findings suggest that data retrieval can boost performance by selecting relational data and filtering noisy data. Another grand challenge for data lake systems is supporting various on-demand queries. Our \sys offers a combination of data modules and task modules in a flexible way. From a data module perspective, automatic data retrieval is used to extract useful information, while data parsing is used for data interpretation. From a task module perspective, prompt engineering demonstrates remarkable cross-task capabilities across various data tasks. Overall, our \sys makes data in data lakes actionable and enriches data lakes in a flexible manner.



\section{Conclusion}
\label{sec:conclusion}

In this paper, we design \sys, a unified framework to solve data manipulation tasks on data lakes. \sys summarizes a number of data manipulation tasks into a unified form and designs general steps to solve these tasks using LLMs with proper prompts. Experimental results demonstrate that \sys exhibits superior performance when compared to traditional and learned methods on a variety of data manipulation tasks. In the future work, we show that there still reserves enough room for improvement in terms of data, model and algorithm. We hope the strategies proposed in \sys could contribute and inspire more advances on exploring LLMs with database systems.

\subsection{future works}
\sys can benefit from explicitly retrieving data, potentially containing evidence and factual information, before the processing of data manipulation. However, the model may be unreliable when dealing with domain-specific knowledge that is required for commercial use cases. In such scenarios, LLMs can generate information that may be problematic in practical applications where factual accuracy is crucial. Another limitation is explainability. In most data applications (e.g., root cause analysis), a good explanation that includes all the rules applied to reach a conclusion can be valuable to the user. While LLMs have exhibited remarkable success in various data tasks, their ability to reason and explain is often viewed as a limitation. We summarize some future research directions in terms of different perspectives of data, model, algorithm and efficiency as follows.

\sstitle{Integration with Domain Knowledge}
From the results in Section~\ref{sec:exp-2}, we observe that \sys can perform well on universal data but may fall on domain specific data. However, data lakes often contain data from highly specialized domains, e.g., financial, biological and academic data. 
Currently, the widely adopted method is to fine-tune the LLMs with domain specific data. Whereas, there still remain challenges for fine-tuning, such as how to extract high quality data from data lakes as a corpus to tune the LLMs.
Besides, it is very interesting to explore new integration methods except fine-tuning LLMs.


\sstitle{Designing Large Models for DB Tasks}
The LLMs are mainly trained on a corpus of texts to resolve NLP tasks. Although we could design serialization functions and prompts to apply LLMs on tabular data, it is essentially a process of cutting the foot to fit the shoe. A better way is to design and train large models on tabular (and other types of) data from scratch to capture semantics for database tasks. Some previous works have attempted to leverage BERT (e.g., TaPas \cite{herzig2020tapas}, TaPEx \cite{liu2021tapex}, Tableformer \cite{yang2022tableformer}, TURL\cite{deng2022turl}) to understand (semi-)structured data. However, all of the concepts, model structures, training methods and the whole paradigm of large models need to be re-designed to fit database tasks.

\sstitle{Efficiency Consideration}
LLMs applied in our method bring benefits but also entail an increase in computational resource. In the future work, it is rather important to consider how to improve the efficiency while retaining the effectiveness of LLM-based methods. One possible way is to design more efficient retrieval methods to extract relevant information from data to minimize the computation overhead. Another way is to adapt to select the LLMs with minimal computation cost to fulfill each task. As we show in Table~\ref{tab:ft}, a fine-tuned LLM in a smaller size is possible to match the performance of a universal LLM in a much larger size. 

\sstitle{LLM-based and Traditional Methods}
In spite of our LLM-based solutions exhibiting superiority in terms of result effectiveness, they can not totally replace traditional methods. LLM is still a black-box which is difficult to interpret, debug and analyze. These are all risky factors for database systems which require rock-solid stability. Traditional methods relying on rules and logic tuned by human experience over decades have their unique advantages. They are more friendly to system deployment. Therefore, LLM-based and traditional methods are not conflicting but rather complementary to each other. It would be very practical to combine their advantages together to control the deployment risk while still attaining high result effectiveness.


\section*{Acknowledgements}
The work was supported by grants from the National Natural Science Foundation of China (No. 92367110) and  the program of Alibaba Innovative Research.

\clearpage


\bibliography{example_paper}
\bibliographystyle{mlsys2024}

\clearpage


\appendix
\appendix
\appendixpage

\eat{
\section*{Appendix overview}
\label{app}

In the appendix, we first provide more complex tasks supported by our method, including TabelQA (Appendix \ref{app:sec-qa}), join discovery (Appendix \ref{app:sec-jd}), and information extraction (Appendix \ref{app:sec-ie}). In Appendix \ref{app:sec-algo}, we also represent the main function, executed by \sys, in pseudo-code. 
}

\section{Prompt Template}
\label{app: sec-pt}

We provide a running example to explain how our method automatically generates the desired cloze question of target task based on the in-context learning of LLMs. The prompt example is as follows:

\noindent\fbox{\parbox{0.98\linewidth}{
\colorbox{mygray}
{
\begin{minipage}{0.95\linewidth}

\scriptsize{
\vspace{2.0em}
\textbf{(Input to LLMs):\\}
\\
Write the claim as a cloze question.

\textbf{Claim:}\\
The task is data imputation which produces the missing data with some value to retain most of the data. The context is Wenham, Marysville, and Westmont are cities in the United States, identified by the ISO3 code USA. The target is city:New Cassel, iso3:USA, country:?\\
\textbf{Cloze question:}\\
Wenham, Marysville, and Westmont are cities in the United States, identified by the ISO3 code USA.
New Cassel is the name of a city whose ISO3 country code is USA. New Kassel belongs to the country \_.

\textbf{Claim:}\\
The task is data transformation which is the process of converting data from one format to another required format within a record. The context is data before transformation: 20000101 data after transformation: 2000-01-01. The target is 19990415:?\\
\textbf{Cloze question:}\\
20000101 can be transformed to 2000-01-01, and 19990415 can be transformed to \_.

\textbf{Claim:}\\
The task is error detection which detect attribute error within a record in a data cleaning system. The context is the address of 2505 u s highway 431 north is not an error, the county name of mxrshxll is an error. The target is whether there is an error in city:sheffxeld.\\
\textbf{Cloze question:}\\
The address "2505 U.S. Highway 431 North" has no error, whereas the county name "mxrshxll" contains an error. It is required to identify if there is an error in the city name "sheffxeld". Is there an error in the city name? Yes or No. \_

\textbf{Claim:}\\
The task is entity resolution which is the process of predicting whether two records are referencing the same real-world thing. The context is A is the Punch! Home Design Architectural Series 4000 v10, manufactured by Punch! Software, is priced at \$199.99. B is The Punch Software 41100 Punch! Home Design Architectural Series 18, manufactured by Punch Software, is priced at \$18.99. The target is are A and B the same?\\
\textbf{Cloze question:}\\

\textbf{(Output of LLMs):\\}
Punch! Home Design Architectural Series 4000 v10, manufactured by Punch! Software, is priced at \$199.99, whereas Punch Software 41100 Punch! Home Design Architectural Series 18, also manufactured by Punch Software, is priced at \$18.99. Are these two products the same? Yes or No. \_
\vspace{2.0em}
}

\end{minipage}
}
}}

\section{Case Study}
\label{app: sec-cs}

For case study, we present the final results of FM (random setting), FM (manual setting), and our retrieval-based UniDM. For FM method (random setting), we have the final prompt and the output of LLMs as follows:

\noindent\fbox{\parbox{0.98\linewidth}{
\colorbox{mygray}
{
\begin{minipage}{0.95\linewidth}

\scriptsize{
\vspace{2.0em}
\textbf{(Input to LLMs):\\}
name: anthonys. addr: 3109 piedmont rd.  just south of peachtree rd.. phone: 404/262-7379. type: american. What is the city? atlanta\\
name: rose pistola. addr: 532 columbus ave.. phone: 415/399-0499. type: italian. What is the city? san francisco\\
name: american place. addr: 2 park ave. at 32nd st.. phone: 212/684-2122. type: american. What is the city? new york\\
name: ruth's chris steak house (los angeles). addr: 224 s. beverly dr.. phone: 310-859-8744. type: steakhouses. What is the city?\\
\\
\textbf{(Output of LLMs)}: \textcolor{red}{los angeles}\\
\textbf{Ground Truth}: Beverly Hills
\vspace{2.0em}
}

\end{minipage}
}
}}

For FM method (manual setting), we have the final prompt and the output of LLMs as follows:

\noindent\fbox{\parbox{0.98\linewidth}{
\colorbox{mygray}
{
\begin{minipage}{0.95\linewidth}

\scriptsize{
\vspace{2.0em}
\textbf{(Input to LLMs):\\}
name: oceana. addr: 55 e. 54th st.. phone: 212/759-5941. type: seafood. What is the city? new york\\
name: oceana. addr: 55 e. 54th st.. phone: 212-759-5941. type: seafood. What is the city? new york city\\
name: ruth's chris steak house (los angeles). addr: 224 s. beverly dr.. phone: 310-859-8744. type: steakhouses. What is the city?\\
\\
\textbf{(Output of LLMs)}: \textcolor{red}{los angeles}\\
\textbf{Ground Truth}: Beverly Hills
\vspace{2.0em}
}
\end{minipage}
}
}}

For our retrieval-based \sys, we have the final prompt and the output of LLMs as follows:

\noindent\fbox{\parbox{0.98\linewidth}{
\colorbox{mygray}
{
\begin{minipage}{0.95\linewidth}

\scriptsize{
\vspace{2.0em}
\textbf{(Input to LLMs):\\}
The name of the place is Belvedere. The address is 9882 Little Santa Monica Blvd. The city is Beverly Hills.\\
The name of the grill is Jack Sprat's Grill and its address is 10668 W. Pico Blvd. in the city of Los Angeles.\\
The name of the establishment is Border Grill, located on 4th Street in Los Angeles.\\
Ruth's Chris Steak House (Los Angeles) is located at 224 S. Beverly Dr. Ruth's Chris Steak House (Los Angeles) is located in the city of \_.\\
\\
\textbf{(Output of LLMs)}: \textcolor{orange}{Beverly Hills}\\
\textbf{Ground Truth}: Beverly Hills
\vspace{2.0em}
}
\end{minipage}
}
}}

\clearpage
\color{black}

\section{Explaining the Table Question Answer Task}
\label{app:sec-qa}

\begin{figure}[ht]
    \includegraphics[width=\linewidth]{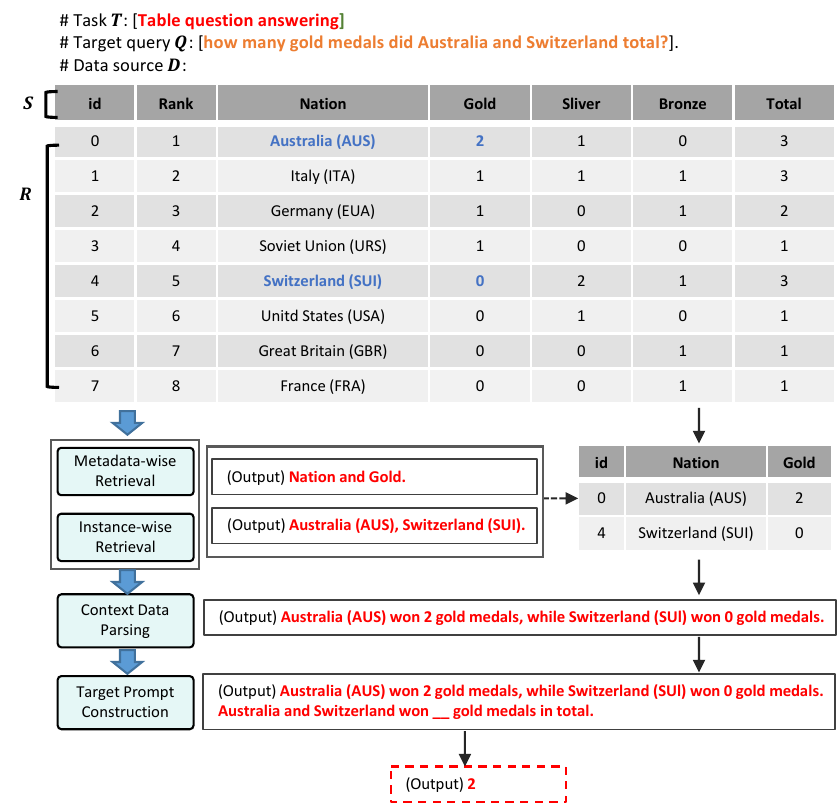}
\vspace{-1.5em}
        \caption{Explanation example for table question answering by the \sys.}
\vspace{-1.0em}
\label{fig:TableQA}    
\end{figure}

To show the generality of our \sys solution, we apply that it could be applied to the more complex table question answer (TableQA) task. This is a task to ask a question to retrieve answers from a data table. \autoref{fig:TableQA} gives an illustrative example on WikiTableQuestions dataset~\cite{wikiqa2015}. Here we have a question $Q$: 
``how many gold medals did Australia and Switzerland total?'' and the answer on the number of gold medals could be obtained from the table by finding the relevant information. 

For the TableQA task, we directly set the task query $Q$ to be the question contained in the task description. The set $R$ of records and set $S$ of attributes are set to contain all records and attributes in the table $D_i$, respectively. When applying \sys to solve TableQA, in the first context information retrieval, we set the set of candidate attributes $S' = S$. By applying prompts $p_{rm}$ and $p_{ri}$, \sys first automatically retrieves a content snapshot $\mathcal{C}$ from the data table $D_i$. This snapshot contains a selection of columns (`Nation' and `Gold' in our example) and rows (`Australia (AUS)' and `Switzerland (SUI)' in our example) that summarize the information most relevant to the task and the query (`how many \underline{gold medals} did \underline{Australia} and \underline{Switzerland} total?').

In the second step, the context snapshot $\mathcal{C}$ is serialized and then parsed into a natural text representation $\mathcal{C}'$. In our example, we now know that ``\underline{Australia (AUS)} won \underline{2} goal medals, while \underline{Switzerland (SUI)} won \underline{0} goal medals''. 
To facilitate open-ended cloze question generation centered around the target query and contextual information drawn from the data, the prompt engineering module is employed. Ultimately, the resulting cloze question is fed into the LLM to yield an answer. \sys could correctly output ``2'' as the answer. 
This exhibit that \sys is good at not only processing instance-level tasks such as data imputation and error correction, but also can be applied to retrieve table-level information.

\begin{figure}[!tbp]
  \centering
  \includegraphics[width=\linewidth]{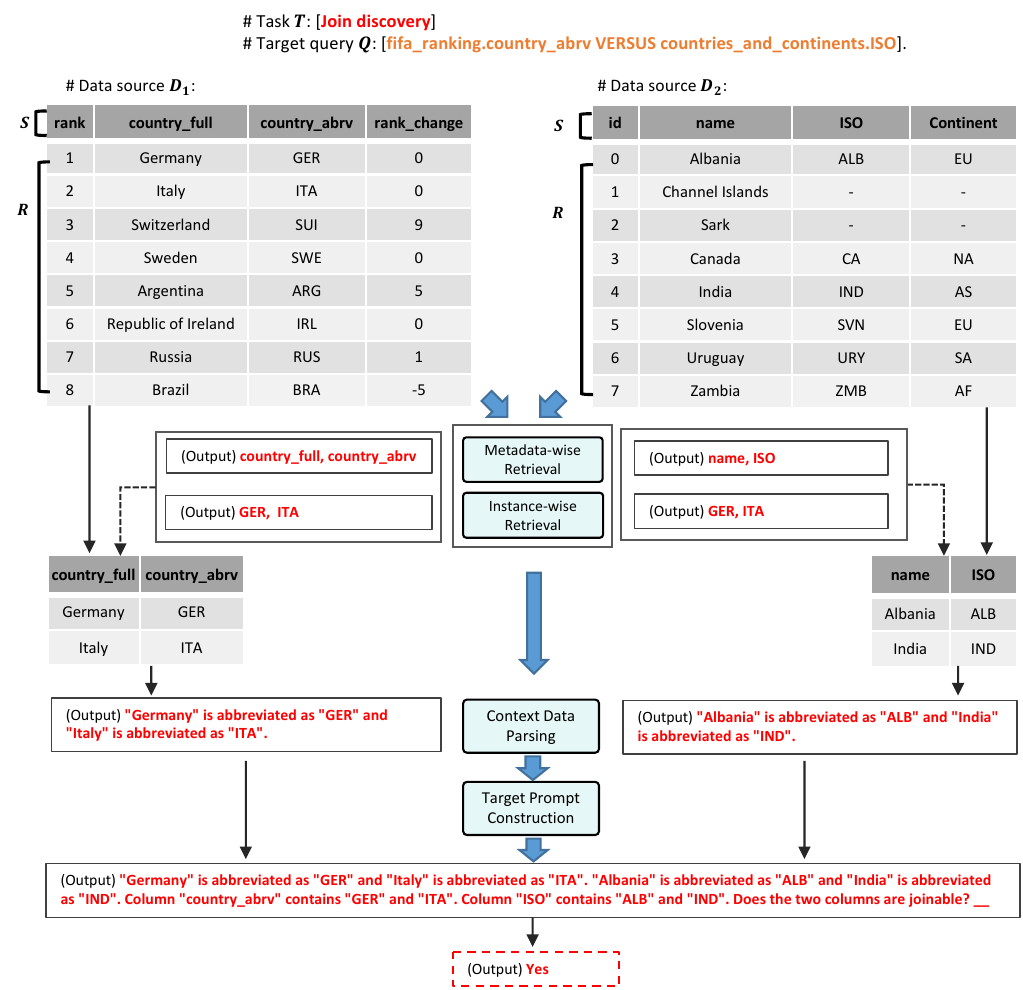}
\vspace{-1.5em}
      \caption{Explanation example for join discovery by the \sys.} 

\label{fig:exam_join}
\end{figure}

\begin{figure}[!tbp]
  \centering
  \includegraphics[width=0.95\linewidth]{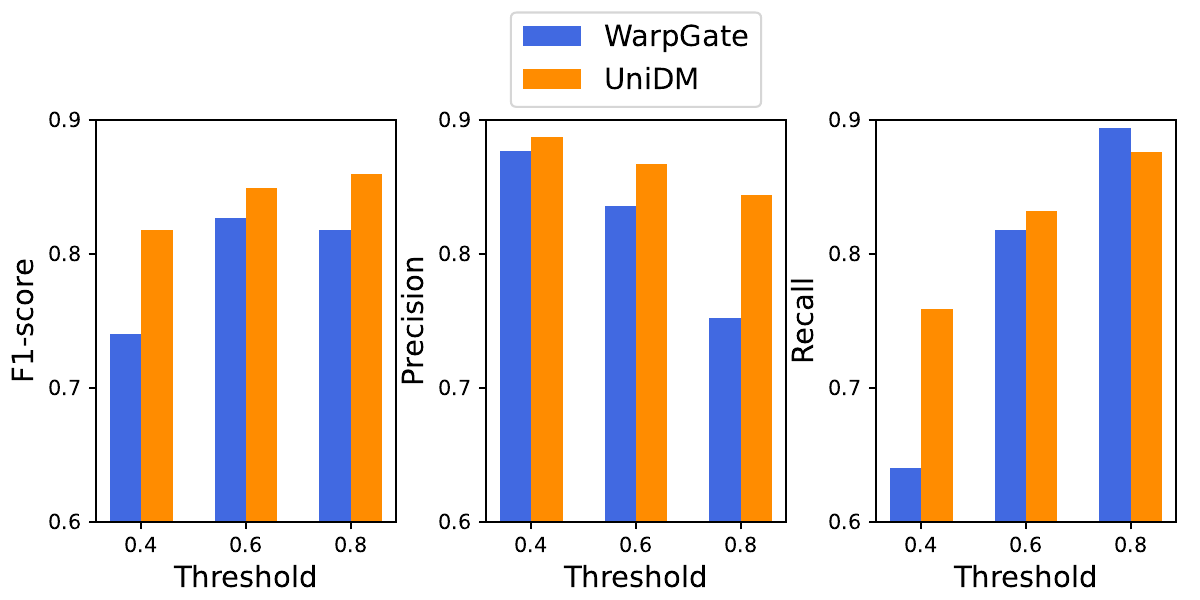}
\vspace{-1.0em}
      \caption{F1-score, precision and recall on join discovery task.} 
\vspace{-1.0em}
\label{fig:join}
\end{figure}

\section{Join Discovery Task}
\label{app:sec-jd}

In data lakes, the join relations across tables are not specified. Join discovery task is a major challenge in data analysis. It aims at finding semantically joinable columns across different tables. This task could be subsumed and solved by our \sys framework. 

\eat{
First, we show how \sys subsumes the join discovery task. Let $S \subseteq S_i$ be a singleton attribute in $S_i$ of table $D_i \in \mathcal{D}$. $R$ contains all records in the target table $D_i$ where $i \neq j$. The function $F_{T}(R, S, \mathcal{D})$ outputs all attributes in $S_j$ that could be joined with attributes in $S_i$.

To solve this task by \sys, we set the task query parameter $Q = S_i$ to show the target attributes. In the context information retrieval step, we set the set of candidate attributes $S' = \cup_{j \neq i} S_j$ to contain all attributes in other tables. \sys retrieves some representative records under attributes in $S'$ and then 
}

\autoref{fig:exam_join} gives an illustrative example of join discovery task. The query $Q$ is set to be the textual name of the two tables, e.g., ``fifa\_ranking.country\_abrv '' and ``countries\_and\_continents.ISO''. The set $R$ of records and the set $S$ of attributes are set to contain all records and attributes in the two tables $D_1$ and $D_2$, respectively. In the first context information retrieval, \sys extracts joinable attributes and records $\mathcal{C}_1$ and $\mathcal{C}_2$ between the two tables. The context snapshots $\mathcal{C}_1$ and $\mathcal{C}_2$ are serialized and then parsed into natural text representations $\mathcal{C}'_1$ and $\mathcal{C}'_{2}$ separately. 
By using the prompt engineering module, \sys constructs a cloze question with the target query and contextual information from the two tables.
Ultimately, the resulting cloze question is fed into the LLM to yield an answer ``Yes'' that indicates the two tables are joinable.

For experiment on join discovery task, we use NextiaJD \citep{flores2021towards} that composes four splits according to their file size. The dataset labels the join quality of attribute pairs based on a measure that considers both containment and cardinality proportion with empirically determined thresholds. In experiments, we use a subset with 4404 pairs (2239 positive and 2164 negative) of attributes whose quality is labeled as Good and High. For the baseline, we select an embedding-based solution WarpGate \citep{cong2022warpgate}, the SOTA method. As shown in \autoref{fig:join}, we report precision, recall and F1-score under various threshold values. We find that \sys consistently obtains higher F1-score compared with WarpGate \citep{cong2022warpgate} under various threshold values. This exhibits that UniDM's potential to manipulate data cross sources. 

\section{Information Extraction Task}
\label{app:sec-ie}

\begin{figure}[ht]
  \centering
  \includegraphics[width=\linewidth]{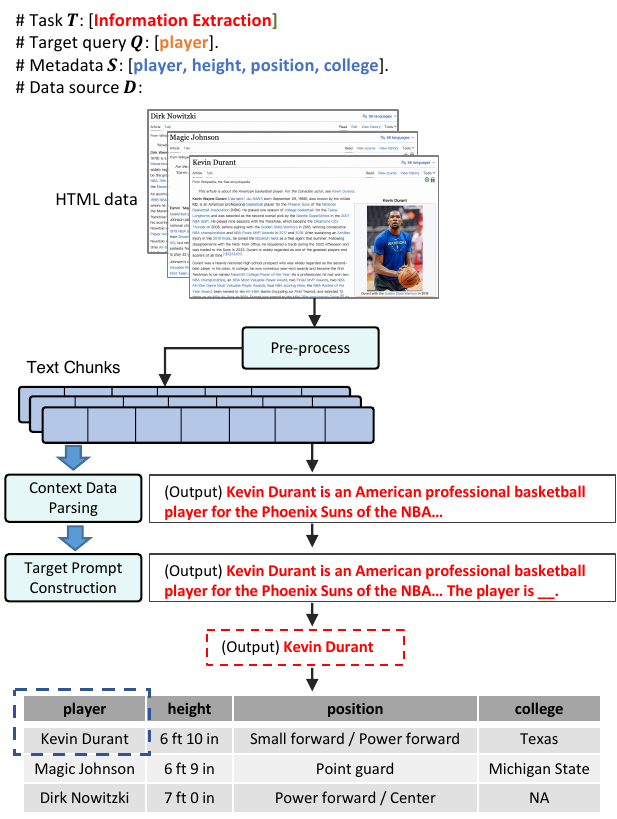}
\vspace{-1.5em}
      \caption{Explanation example for information extraction by the \sys.} 
\vspace{-1.0em}
\label{fig:ie}
\end{figure}

\begin{table}[ht]
\caption{Text F1-score on information extraction task.}
\label{tab:ie}
\begin{tabular}{c|c}
\multirow{2}{*}{\textbf{Method}} & \textbf{Information Extraction} \\
                                 & \textbf{NBA player}             \\ 
\thickhline
Evaporate-code                   & 40.6                            \\
Evaporate-code+                  & 84.6                            \\
\sys                            & 70.1                            
\end{tabular}
\end{table}

In order to demonstrate the performance of \sys in handling more complex data manipulation tasks, we conduct experiments on an information extraction task. For the information extraction task, we aim to construct a structured view (i.e., tabular) of a set of semi-structured documents (e.g., HTML). 

\autoref{fig:ie} gives an illustrative example of information extraction task. $D_i$ is a semi-structured document and $R=D_i$. $S$ represents a set of pre-defined attributes to be extracted from $D_i$. $F_{T}(R, S, \mathcal{D})$ outputs a set of attribute values from the document $D_i$. The query $Q$ is set to be the target attribute $S$, e.g., ``player''. Note that we temporarily removed the context retrieval module because the attributes and the instance is pre-defined by users. We slightly modified the context parsing prompt by adding the query $Q$. It aims to extract and transform the context information (``Kevin Durant is an American professional basketball player...'') according to the query $Q$. We then construct the target cloze question and yield the final result (i.e., ``Kevin Durant'').

For information extraction, we use SWDE benchmark in~\cite{hao2011from} and choose NBAplayer dataset. This dataset includes complex HTML from the NBA Wikipedia pages for NBA players. Following the previous method, we consult closed information extraction setting, where a pre-defined schema is provided and \sys is used to populate the table. Ground truth value is available for the target attributes.  
As shown in \autoref{tab:ie}, \sys outperforms the baseline methods Evaporate-code in terms of the F1-score. Evaporate-code$+$ achieves better results due to its utilization of ensemble methods. This exhibits that \sys's potential to manipulate semi-structured data. 

\section{Algorithm}
\label{app:sec-algo}

Algorithm \autoref{alg:1} represents the main function for data manipulation tasks, executed by \sys, in pseudo-code. The input integrates a data lake for a target data manipulation task, with user-provided parameters. These parameters include a subset of schema, a subset of records, a task description, and a target query. First, we automatically retrieve context information according to the task and the input query. This module uses LLMs to select valuable attributes for the task and the target attribute, and then perform a fine-grained filtering on all records to identify relevant ones w.r.t. target records.
Next, the context information $\mathcal{C}$, represented in a tabular form, is transformed into a more effective format $C'$ for LLMs. The target prompt construction is to find an effective prompt to organize the task description $T$, the context information $C'$ and the query $Q$. Finally, this prompt is fed into LLMs to yield the final answer of our task.

\begin{algorithm}[hb]
    \caption{Unified Framework for Data Manipulation (\sys)}
    \label{alg:1}
    \renewcommand{\algorithmicrequire}{\textbf{Input:}}  
 	\begin{algorithmic}[1]
		\REQUIRE a data lake $\mathcal{D}$, a subset of records $R \subseteq D_i$ extracted from a table $D_i \in \mathcal{D}$, a subset of attributes $S \subseteq S_i$ under the schema $S_i$, a task description $T$, a target query $Q$.
        \IF{Context Retrieval}
            \STATE // Metadata-wise retrieve
            \STATE $p_{rm} \gets \text{prompt\_}p_{rm}(T,Q,S)$
            \STATE $S^m \gets \text{LLM}(p_{rm})$
            \STATE // Instance-wise retrieve
            \STATE $p_{ri} \gets \text{prompt\_}p_{ri}(T,Q,R)$
            \STATE $\{score_i\}^m_{i=1} \gets \text{LLM}(p_{ri})$
            \STATE $R^m \gets \text{top-k}(\{score_i\})$
            \STATE // Select records based on the retrieved results
            \STATE $\mathcal{C} \gets R^m[S^m]$
        \ELSE{}
            \STATE $\mathcal{C} \gets D_i$
        \ENDIF
        \STATE $\text{serialize}(C) = \{s:r[s]\}, (\forall r[s] \in C)$
        \STATE $\mathcal{V} \gets \text{serialize}(C)$
        \IF{Context Data Parsing}
            \STATE // Parse the data into a natural text representation
            \STATE $p_{dp} \gets \text{prompt\_}p_{dp}(\mathcal{V})$
            \STATE $\mathcal{C}^\prime \gets \text{LLM}(p_{dp})$
        \ELSE{}
            \STATE $\mathcal{C}^\prime \gets \mathcal{V}$
        \ENDIF 
        \STATE // Recursively uses the LLM to reformat the data task.
        \STATE $p_{cq} \gets \text{prompt\_}p_{cq}(T,Q,C^\prime)$
        \STATE $p_{tg} \gets \text{LLM}(p_{cq})$
        \STATE $Y \gets \text{LLM}(p_{tg})$
        \STATE Return $Y$
\end{algorithmic}
\end{algorithm}





\end{document}